# Adaptive mixed norm optical flow estimation

Vania Vieira Estrela [a*], Matthias O. Franz [b], Ricardo T. Lopes [c], A. P. De Araújo [a]

[a] LCMAT-CCT, Grupo de Computação Científica Universidade Estadual do Norte Fluminense (UENF), Av. Alberto Lamego 2000, CEP 28013-600, Campos dos Goytacazes, RJ, BRAZIL;
[b] Max Planck Institute for Biological Cybernetics Spemannstraße 38, 72076 Tübingen, GERMANY;
[c] Nuclear Instrumentation Laboratory/EE, COPPE/UFRJ, CEP 21945-970, RJ, BRAZIL

**ABSTRACT**

The pel-recursive computation of 2-D optical flow has been extensively studied in computer vision to estimate motion from image sequences, but it still raises a wealth of issues, such as the treatment of outliers, motion discontinuities and occlusion. It relies on spatio-temporal brightness variations due to motion. Our proposed adaptive regularized approach deals with these issues within a common framework. It relies on the use of a data-driven technique called Mixed Norm (MN) to estimate the best motion vector for a given pixel. In our model, various types of noise can be handled, representing different sources of error. The motion vector estimation takes into consideration local image properties and it results from the minimization of a mixed norm functional with a regularization parameter depending on the kurtosis. This parameter determines the relative importance of the fourth norm and makes the functional convex. The main advantage of the developed procedure is that no knowledge of the noise distribution is necessary. Experiments indicate that this approach provides robust estimates of the optical flow.

## 1. INTRODUCTION

*Motion detection* is used to determine if motion is present between two video frames and the result is usually a region of interest indicating where motion occurred. *Motion estimation* is the process of identifying motion between two frames within a region of interest. In this case, the output is normally the magnitude and direction of the motion, either as a whole or on a per-pixel basis (pel-recursive approach). Pel-recursive algorithms can also manage motion with sub-pixel accuracy.

In video coding, the estimated motion can be used to reduce the transmission bandwidth. The evolution of an image sequence motion field can also help other image processing tasks in multimedia applications such as analysis, recognition, tracking, restoration, collision avoidance, segmentation of objects, surveillance, virtual reality, and navigation. Autonomous mobile robots represent an interesting, but extremely challenging application of motion estimation

Motion compensation is an important concept for efficient transmission of time-varying images. The similarities between successive frames of a video signal are exploited in order to achieve a good data compression.

In coding applications, a block-based approach is often used for interpolation of lost information between key frames[8,12]. The fixed rectangular partitioning of the image used by some block-based approaches often separates visually meaningful image features. Pel-recursive schemes[1,2,3,8] can theoretically overcome some of the limitations associated with blocks by assigning a unique motion vector to each pixel. Intermediate frames are then constructed by resampling the image at locations determined by linear interpolation of the motion vectors. The pel-recursive approach can also manage motion with sub-pixel accuracy. The update of the motion estimate was based on the minimization of the displaced frame difference (DFD) at a pixel. In the absence of additional assumptions about the pixel motion, this estimation problem becomes "ill-posed" because of the following problems: a) occlusion; b) the solution to the 2D motion estimation problem is not unique (aperture problem); and c) the solution does not continuously depend on the data due to the fact that motion estimation is highly sensitive to the presence of observation noise in video images.

We propose to solve optical flow (OF) problems by means of a mixed norm framework. Such approach accounts better for the statistical properties of the errors present in the scenes than the solution proposed in previous works[1,2,3] relying on the assumption that the contaminating noise has Gaussian distribution. A more realistic approach is to take into consideration non-Gaussian noise distributions. This can be accomplished by means of higher order norms such as what is done in the least mean fourth (LMF) algorithm[5,7,9]. The kurtosis of a random signal is a measure of its Gaussianity. The mixed-norm parameter is a function of the noise kurtosis and it can obtained at each step from the available data. It is important to highlight the fact that, with the exception of the "smoothness" assumption, no prior knowledge regarding the properties of the image and the noise is necessary.

The reminder of the work is organized as follows: Section 2 provides some necessary background on the pel-recursive motion estimation problem. Section 3 introduces the concept of kurtosis of a random signal. Section 4 describes our proposed technique and some details of the computer implementation of the given method. Section 5 describes the experiments used to access the performance of our proposed algorithm. Finally, the main conclusions are summarized in Section 6.

## 2. PROBLEM FORMULATION

The displacement of each picture element in each frame forms the displacement vector field (DVF) and its estimation can be done using at least two successive frames. The DVF is the 2-D motion resulting from the apparent motion of the image brightness (OF). A vector is assigned to each point in the image.

A pixel belongs to a moving area if its intensity has changed between consecutive frames. Hence, our goal is to find the corresponding intensity value $I_k(\boldsymbol{r})$ of the k-th frame at location $\boldsymbol{r} = [x, y]^T$, and $\boldsymbol{d}(\boldsymbol{r}) = [d_x, d_y]^T$ the corresponding (true) displacement vector (DV) at the working point $\boldsymbol{r}$ in the current frame. Pel-recursive algorithms minimize the DFD function in a small area containing the working point assuming constant image intensity along the motion trajectory. The DFD is defined by

$$\Delta(\boldsymbol{r}; \boldsymbol{d}(\boldsymbol{r})) = I_k(\boldsymbol{r}) - I_{k-1}(\boldsymbol{r}-\boldsymbol{d}(\boldsymbol{r})),$$

and the perfect registration of frames will result in $I_k(\boldsymbol{r})=I_{k-1}(\boldsymbol{r}-\boldsymbol{d}(\boldsymbol{r}))$. The DFD represents the error due to the nonlinear temporal prediction of the intensity field through the DV. The relationship between the DVF and the intensity field is nonlinear. An estimate of $\boldsymbol{d}(\boldsymbol{r})$, is obtained by directly minimizing $\Delta(\boldsymbol{r},\boldsymbol{d}(\boldsymbol{r}))$ or by determining a linear relationship between these two variables through some model. This is accomplished by using the Taylor series expansion of $I_{k-1}(\boldsymbol{r}-\boldsymbol{d}(\boldsymbol{r}))$ about the location $(\boldsymbol{r}-\boldsymbol{d}^i(\boldsymbol{r}))$, where $\boldsymbol{d}^i(\boldsymbol{r})$ represents a prediction of $\boldsymbol{d}(\boldsymbol{r})$ in $i$-th step. This results in

$$\Delta(\boldsymbol{r}, \boldsymbol{r}-\boldsymbol{d}^i(\boldsymbol{r})) = - \boldsymbol{u}^T \nabla I_{k-1}(\boldsymbol{r}-\boldsymbol{d}^i(\boldsymbol{r})) + e(\boldsymbol{r}, \boldsymbol{d}(\boldsymbol{r})), \qquad (1)$$

where the displacement update vector $\boldsymbol{u}=[u_x, u_y]^T = \boldsymbol{d}(\boldsymbol{r}) - \boldsymbol{d}^i(\boldsymbol{r})$, $e(\boldsymbol{r}, \boldsymbol{d}(\boldsymbol{r}))$ represents the error resulting from the truncation of the higher order terms (linearization error) and $\nabla=[\partial/\partial_x, \partial/\partial_y]^T$ represents the spatial gradient operator. Applying Eq. (1) to all points in a neighborhood $\mathcal{R}$ gives

$$\boldsymbol{z} = \boldsymbol{G}\boldsymbol{u} + \boldsymbol{n}, \qquad (2)$$

where the temporal gradients $\Delta(\boldsymbol{r}, \boldsymbol{r}-\boldsymbol{d}^i(\boldsymbol{r}))$ have been stacked to form the $N\times 1$ observation vector $\boldsymbol{z}$ containing DFD information on all the pixels in a neighborhood $\mathcal{R}$, the $N\times 2$ matrix $\boldsymbol{G}$ is obtained by stacking the spatial gradient operators at each observation, and the error terms have formed the $N\times 1$ noise vector $\boldsymbol{n}$ which is assumed Gaussian with $\boldsymbol{n}\sim N(0, \sigma_n^2 \boldsymbol{I})$. Each row of $\boldsymbol{G}$ has entries $\boldsymbol{g}_i = [g_{xi}, g_{yi}]^T$, with $i = 1, \ldots, N$. The spatial gradients of $I_{k-1}$ are calculated through a bilinear interpolation scheme[2,3].

## 3. LOCALIZED KURTOSIS

In this section, a statistical framework is introduced as a means to circumvent some difficulties related to the existence of noise whose probability density function (pdf) is not known *a priori*. In most applications the contaminating noise is assumed to be zero-mean Gaussian. Nevertheless, there are applications for which the additive noise is characterized by other probability distributions such as uniform, Laplacian or by a combination of several noise types.

The non-Gaussianity of the optical flow in a neighborhood can lead to inconsistent motion estimates. The kurtosis is a way to determine the flatness of the pdf of a signal. For a zero-mean random variable *n*, its kurtosis can be used to evaluate its Gaussianity and it is defined by

$$\chi(n) = E\{n^4\} - 3E^2\{n^2\},$$

where E{.} stands for the expectation operator. Normally distributed signals have kurtosis equal to zero. Super-Gaussian or leptokurtic signals such the ones with Laplacian distributions and sub-Gaussian or platykurtic signals have, respectively, positive and negative values of χ(*n*).

An image sequence can be contaminated by two or more combined noise distributions. Since the kurtosis of the linear combination of uncorrelated zero-mean types of noise $n_1$ and $n_2$ is

$$\chi(an_1 + bn_2) = a\chi(n_1) + a\chi(n_2),$$

where *a* and *b* are positive scalars.

## 4. MIXED-NORM (MN) ALGORITHM

A solution to Eq. (2) can be obtained by minimizing the following convex mixed-norm functional of the kurtosis:

$$J(u) = [1 - \gamma(n)]\|z - Gu\|_2^2 + \gamma(n)\|z - Gu\|_4^4, \tag{3}$$

where the mixed-norm parameter γ(*n*) ∈ [0, 1] controls the relative importance of the second and fourth norms similarly to the way one introduces regularization in a linear model such as proposed elsewhere[4,10]. *J*(*u*) is called the least mixed-norm functional and all signals are ergodic. When the kurtosis is zero, the contribution of the second order term becomes greater as *A* increases. At each iteration, a positive kurtosis means that the importance of the forth norm is decreased when it is compared to the second norm. If the kurtosis is negative, then the forth norm dominates the second norm.

The performance of the least mean square (LMS) and the LMF algorithms have been investigated in the literature. It was shown that under certain noise conditions, for instance with sub-Gaussian signals, the LMF and other higher order criteria exhibit improved performance when compared to the LMS. The converse is true for Gaussian and super-Gaussian signals. According to Eq. (3), it is desired that in the extreme cases of only Gaussian or super-Gaussian noise, the contribution of the fourth norm is negligible (γ(*n*)≈0), while for sub-Gaussian noise the relative contribution of the fourth norm is large (γ(*n*)≈1). On the other hand, for the cases of mixed noise *n*, we will attain values responsible for balancing the relative contribution of the two norms in *J*(*u*). However, for most practical situations, the noise distribution is not known. Hence, we want to determine the values of γ(*n*) using the available data.

We propose to use a steepest descent algorithm for determining the value $u_k$ that minimizes the functional from Eq. (3). First, we will need the gradient of *J*(*u*) with respect to *u* which is equal to

$$\nabla_u J(u) = -2(1 - \gamma(u))G^T(z - Gu) - 4\gamma(u)G^T(z - Gu)^3$$
$$+ [-\nabla_u \gamma(u)\|z - Gu\|_2^2 + \nabla_u \gamma(u)\|z - Gu\|_4^4].$$

A closer look at the last term inside the brackets reveals that it is very small, and therefore it is omitted in the following analysis. It is also confirmed experimentally that the restoration results are indistinguishable with and without the use of the last term.

The problem at hand is the recovery of the update motion vector or the estimation of *u* using the available information. Minimization of the previous expression leads to the following iterative solution[5]:

$$u_{k+1} = u_k + \beta[(1-\gamma(u_k))G^T(z-Gx_k) + 2\gamma(u_k)G^T(z-Gu_k)^3]$$

$$= u_k + \beta G^T[(1-\gamma(u_k))I + 2\gamma(u_k)P(u_k)](z-Gu_k),$$

where $\beta$ is the relaxation parameter which controls the convergence, as well as, the convergence rate and $P(u)$ is an N x N diagonal matrix with diagonal elements $P(u_k)_{i,i} = (z_i - (Gu_k)_i)^2$.

By means of a convergence analysis[5], it was shown that if a system is contaminated by a noise *n*, then one can find the following convex functional

$$\gamma(u) = \frac{exp(-c\chi(n))}{A + exp(-c\chi(n))},$$

where *c* and *A* are positive scalars. If *c* increases then, $\gamma(u)$ tends to the step function. Furthermore, if *A* increases, the function is shifted to the left.

The above equation works very well for mixtures of noise distributions because of the linearity property of the kurtosis[11].

## 5. EXPERIMENTS

The non-Gaussianity of the ergodic optical flow in a neighborhood can lead to inconsistent motion estimates.

For a zero-mean random variable *n*, its kurtosis $\chi(n)$ can be used to evaluate its Gaussianity. Normally distributed signals have kurtosis equal to zero. Super-Gaussian signals such the ones with Laplacian distributions and sub-Gaussian signals have, respectively, positive and negative values of $\chi(n)$. An image sequence can be contaminated by two or more combined noise distributions. Since the kurtosis of a linear combination of uncorrelated zero-mean types of noise $n_1$ and $n_2$ is the linear combination of their individual kurtoses, our model can handle different types of noise.

The noise distribution is not known in most practical situations and we can determine $\gamma(n)$ using the available data.

To evaluate the performance of the proposed algorithm we estimated the kurtosis by means of a casual sliding window with *m* = 5 which contains the current pixel whose coordinates are *r*, as follows:

$$\chi(n) = \frac{\sum_{i=0}^{m-1} I_i^4}{\left[\sum_{i=0}^{m-1} I_i^2\right]^2} - 3.$$

where $I_i$ is the image intensity of the *i*-th pixel and *m* is the size in pixels of the window.

Figs. 1, 2, 3, and 4 show preliminary results of our model. In Fig. 1, two motion-compensated frames obtained when the "tennis" sequence is corrupted by Laplacian (super-Gaussian) noise with signal-to-noise-ratios (SNR) of 30 dB and 20 dB respectively are shown. Fig. 2 presents the motion-compensated frames for the "Susie" sequence for SNRs of 30 and 20 dB obtained after corrupting them with Laplacian noise. Figs 3 and 4 deal with Gaussian and uniform noise for respectively the "mother and daughter" and the "foreman" sequences for SNRs of 30 and 20 dB.

The SNR is defined as[6]

$$SNR = 10\log_{10}(\sigma^2/\sigma_c^2),$$

where $\sigma^2$ is the variance of the original image and $\sigma_c^2$ is the variance of the noise corrupted image.

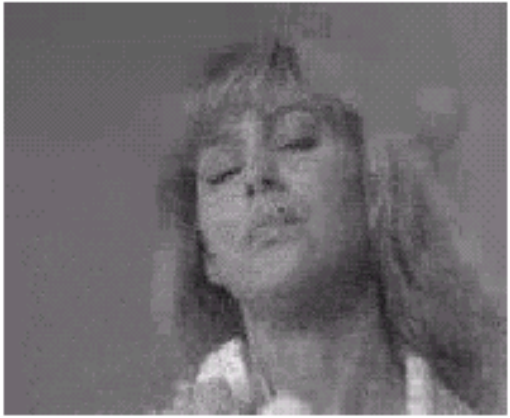 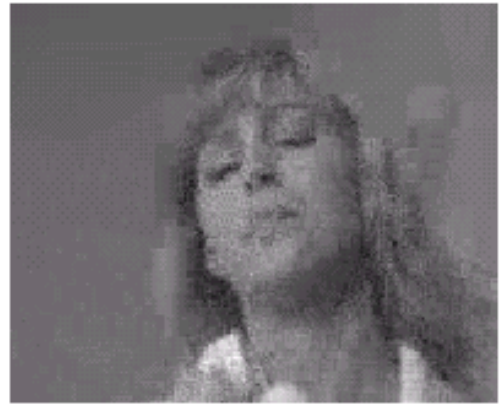

(a) SNR= 30dB  (b) SNR=20dB

**Figure 1.** Motion-compensated frames for the "Susie" sequence (Laplacian noise).

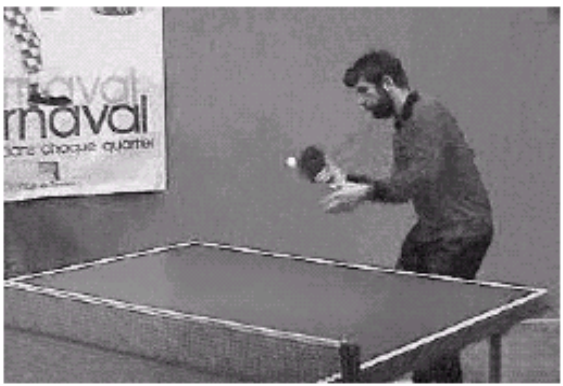 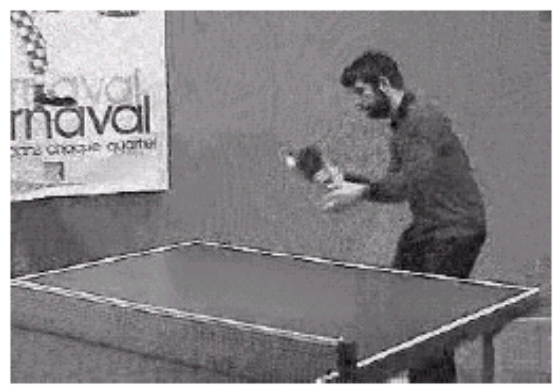

(a) SNR = 30 dB  (b) SNR = 20 dB

**Figure 2.** Examples of motion-compensated frames for the "tennis" sequence (Laplacian noise).

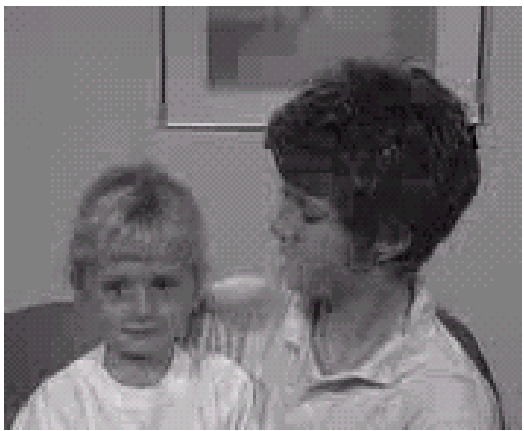 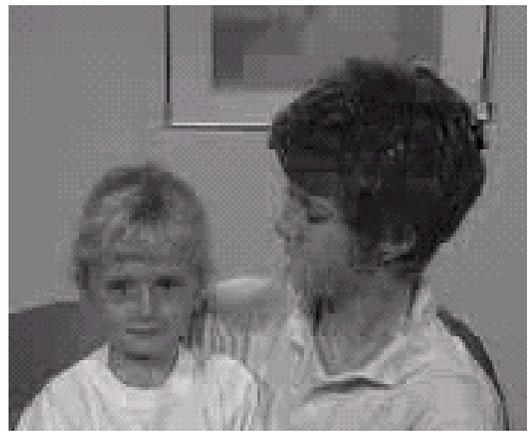

(a) SNR= 30dB  (b) SNR=20dB

**Figure 3.** Examples of motion-compensated frames for the "mother and daughter" sequence (Gaussian noise).

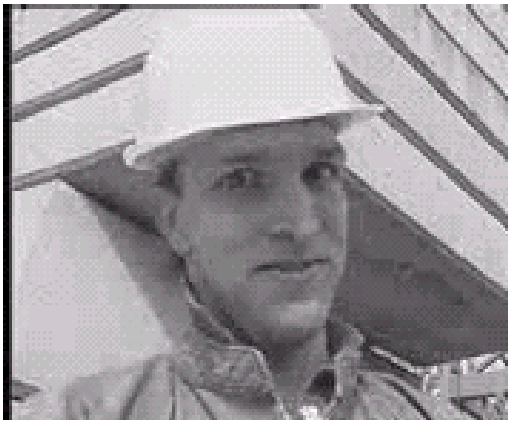 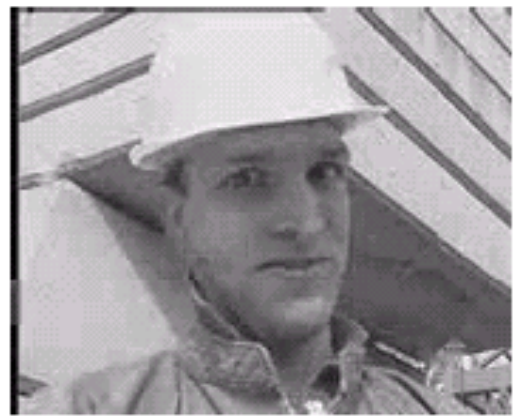

(a) SNR= 30dB  (b) SNR=20dB

**Figure 4.** Examples of motion-compensated frames for the "foreman" sequence (uniform noise).

# 6. CONCLUSIONS

Motion estimation is very important in multimedia video processing applications. Robust statistical approaches are needed because most standard algorithms make unrealistic assumptions about noise distributions which lead to erroneous results that cannot be corrected in subsequent processing stages. The importance of the proposed algorithm relies on the fact that no knowledge of the underlying noise distribution is required, and the relative contribution of the LMS and LMF approaches are adjusted based on the partially motion-compensated image. The mixed norm parameter makes the funtional $J(u)$ convex at each iteration, so that the resulting solution is unique.

An iterative mixed norm model combining the LMS and the LMF functionals and relying on data to estimate the best motion vector for a given pixel was introduced. Some advantages of our model are: a) it handles various types of noise, representing different sources of error because it includes a fourth-order norm; b) the motion vector estimation takes into consideration local image properties such as the existence of contours and the local kurtosis; c) it results from the minimization of a mixed norm functional with a regularization parameter depending on the kurtosis; and d) no knowledge on the noise distribution is necessary. The mixed norm parameter controls the importance of the second and fourth norms, depending on the local information around a pixel and making the functional convex. Experiments indicate that this approach provides estimates which are more robust to noise and this framework is not restricted to Gaussian distributions as in previous works

The method proposed in this work includes a forth order norm, so that the resulting algorithm can handle more general types of contamining noise such as nonzero-mean or nonsymmetric. It also works well for mixtures of noise with different pdfs. The mixed norm functional used in the estimation of motion is more robust to noise and it is not restricted to Gaussian distributions.

## ACKNOWLEDGMENTS

I am thankful for the ongoing support of Prof. Joaquim Teixeira de Assis and his staff who have aided the authors in accomplishing the work presented for their fruitful comments.